\pdfoutput=1

\documentclass[11pt]{article}

\usepackage[final]{acl}

\usepackage{times}
\usepackage{latexsym}
\usepackage[most]{tcolorbox}
\usepackage{hyperref}

\usepackage[T1]{fontenc}

\usepackage[utf8]{inputenc}

\usepackage{microtype}

\usepackage{listings}
\usepackage{color}

\definecolor{dkgreen}{rgb}{0,0.6,0}
\definecolor{gray}{rgb}{0.5,0.5,0.5}
\definecolor{mauve}{rgb}{0.58,0,0.82}

\usepackage{inconsolata}

\usepackage{graphicx}

\usepackage{geometry}
\usepackage{array}
\usepackage{booktabs}
\usepackage{arydshln}
\usepackage{mathtools}

\definecolor{km}{HTML}{FF0000}

\title{\textsc{TinyStyler}: Efficient Few-Shot Text Style Transfer with Authorship Embeddings}

\author{Zachary Horvitz\textsuperscript{\rm 1}, Ajay Patel\textsuperscript{\rm 2}, Kanishk Singh\textsuperscript{\rm 1},  \\ \textbf{ Chris Callison-Burch\textsuperscript{\rm 2}}, \ \textbf{Kathleen McKeown\textsuperscript{\rm 1}}, \textbf{Zhou Yu\textsuperscript{\rm 1}} \\ \textsuperscript{\rm 1}Columbia University, \textsuperscript{\rm 2}University of Pennsylvania
\\ \\
 \texttt{\small zfh2000@columbia.edu}, 
 \texttt{\small ajayp@seas.upenn.edu},
 \texttt{\small ks4038@columbia.edu} \\
 \texttt{\small ccb@seas.upenn.edu}, 
  \texttt{\small kathy@cs.columbia.edu},
 \texttt{\small zy2461@columbia.edu}
  }

\begin{document}
\maketitle
\begin{abstract}

The goal of text style transfer is to transform the style of texts while preserving their original meaning, often with only a few examples of the target style. Existing style transfer methods generally rely on the few-shot capabilities of large language models or on complex controllable text generation approaches that are inefficient and underperform on fluency metrics. We introduce \textsc{TinyStyler}, a lightweight but effective approach, which leverages a small language model (800M params) and pre-trained authorship embeddings to perform efficient, few-shot text style transfer. We evaluate on the challenging task of authorship style transfer and find \textsc{TinyStyler} outperforms strong approaches such as \textsc{Gpt-4}. We also evaluate \textsc{TinyStyler}'s ability to perform text attribute style transfer (formal $\leftrightarrow$ informal) with automatic and human evaluations and find that the approach outperforms recent controllable text generation methods. Our model has been made publicly available \href{https://huggingface.co/tinystyler/tinystyler}{here}.

\end{abstract}

\newcommand{\uarembeddingresultstable}{
    \begin{table*}
    \centering
    \setlength{\tabcolsep}{1.8pt}
    \fontsize{8}{10}\selectfont
    \begin{tabular}{lrrrr|rrrr|rrrr}
    \toprule
    Method ~~~ & \multicolumn{4}{c|}{Random}& \multicolumn{4}{c|}{Single}& \multicolumn{4}{c}{Diverse}\\
           & \multicolumn{1}{c}{\textsc{Away}} & \multicolumn{1}{c}{\textsc{Towards}} & \multicolumn{1}{c}{\textsc{Sim}} & \multicolumn{1}{c|}{\textsc{Joint}} & \multicolumn{1}{c}{\textsc{Away}} & \multicolumn{1}{c}{\textsc{Towards}} & \multicolumn{1}{c}{\textsc{Sim}} & \multicolumn{1}{c|}{\textsc{Joint}} & \multicolumn{1}{c}{\textsc{Away}} & \multicolumn{1}{c}{\textsc{Towards}} & \multicolumn{1}{c}{\textsc{Sim}} & \multicolumn{1}{c}{\textsc{Joint}} \\ \hline
    $\textsc{Copy}_{\textsc{src}}$   & 0.00 & 0.00 & 1.00 & 0.00 & 0.00 & 0.00 & 1.00 & 0.00 & 0.00 & 0.00 & 1.00 & 0.00 \\
    $\textsc{Copy}_{\textsc{tgt}}$    & 1.00 & 1.00 & 0.00 & 0.00 & 1.00 & 1.00 & 0.00 & 0.00 & 1.00 & 1.00 & 0.00 & 0.00 \\ \hline
    \textsc{Capi}   & 0.42 & 0.02 & 0.89 & 0.17 & 0.56 & 0.01 & 0.93 & 0.07 & 0.41 & 0.01 & 0.87 & 0.08 \\
    \textsc{Cont}   & 0.20 & 0.01 & 0.91 & 0.15 & 0.22 & 0.02 & 0.97 & 0.16 & 0.21 & 0.01 & 0.93 & 0.13 \\
    \textsc{Synm}   & 0.23 & 0.02 & 0.92 & 0.17 & 0.22 & 0.01 & 0.95 & 0.10 & 0.17 & 0.01 & 0.91 & 0.07 \\
    \textsc{Punc}   & 0.23 & 0.02 & 0.93 & 0.24 & 0.25 & 0.02 & 0.97 & 0.19 & 0.26 & 0.02 & 0.90 & 0.18 \\
    \textsc{Emoj}   & 0.27 & 0.04 & 0.93 & 0.25 & 0.29 & 0.06 & 0.95 & 0.27 & 0.27 & 0.02 & 0.93 & 0.17 \\ \hline
    $\textsc{Para}_{\textsc{Neu}}$   & 0.78 & 0.01 & 0.58 & 0.05 & 0.91 & 0.01 & 0.60 & 0.05 & 0.87 & 0.05 & 0.53 & 0.18 \\
    $\textsc{Para}_{\textsc{Div}}$   & 0.75 & 0.01 & 0.69 & 0.10 & 0.91 & 0.02 & 0.71 & 0.11 & 0.83 & 0.04 & 0.70 & 0.18 \\
    \textsc{Ling}   & 0.60 & 0.06 & 0.85 & 0.32 & 0.71 & 0.06 & 0.88 & 0.25 & 0.57 & 0.03 & 0.81 & 0.19 \\
    \textsc{Bert}   & 0.22 & 0.02 & 0.72 & 0.13 & 0.29 & 0.01 & 0.69 & 0.10 & 0.30 & 0.01 & 0.60 & 0.08 \\
    $\textsc{Strap}_{p=0.0}$   & 0.98 & 0.02 & 0.16 & 0.05 & 1.00 & 0.00 & 0.23 & 0.00 & 0.97 & 0.02 & 0.22 & 0.04 \\
    $\textsc{Strap}_{p=0.6}$   & 0.99 & 0.02 & 0.08 & 0.04 & 1.00 & 0.00 & 0.13 & 0.01 & 0.97 & 0.02 & 0.11 & 0.03 \\
    $\textsc{Strap}_{p=0.9}$   & 0.99 & 0.02 & 0.05 & 0.03 & 1.00 & 0.00 & 0.08 & 0.00 & 0.97 & 0.01 & 0.05 & 0.01 \\ 
        \textsc{PGuide}$_{\lambda=200}$ & 0.70 & 0.05 & 0.58 & 0.22 & 0.82 & 0.06 & 0.66 & 0.26 & 0.77 & 0.06 & 0.54 & 0.22 \\ 
\textsc{PGuide}$_{\lambda=800}$ & 0.74 & 0.06 & 0.55 & 0.25 & 0.86 & 0.06 & 0.62 & 0.27 & 0.81 & 0.07 & 0.50 & 0.25 \\ 
\textsc{PGuide}$_{\lambda=1500}$ & 0.77 & 0.06 & 0.51 & 0.25 & 0.88 & 0.06 & 0.57 & 0.26 & 0.84 & 0.08 & 0.44 & 0.25 \\ 
\textsc{PGuide}$_{\lambda=2500}$ & 0.80 & 0.07 & 0.46 & 0.25 & 0.90 & 0.06 & 0.51 & 0.23 & 0.86 & 0.08 & 0.38 & 0.24 \\ 
    \hline
    $\textsc{Styll}_{\text{GPT-3}}$  & 0.78 & 0.07 & 0.45 & 0.23 & 0.91 & 0.11 & 0.48 & 0.29 & 0.87 & 0.12 & 0.44 & 0.30 \\
    $\textsc{Styll}_{\text{BLOOM}}$  & 0.70 & 0.11 & 0.54 & 0.34 & 0.86 & 0.16 & 0.57 & 0.40 & 0.76 & 0.12 & 0.58 & 0.36 \\
    \hline

\textsc{Gpt-3.5} & 0.47 & 0.09 & 0.78 & 0.35 & 0.60 & 0.15 & 0.68 & 0.41 & 0.51 & 0.10 & 0.75 & \underline{0.37} \\ 
\textsc{Gpt-4} & 0.76 & 0.09 & 0.71 & 0.33 & 0.85 & 0.08 & 0.72 & 0.31 & 0.83 & 0.07 & 0.68 & 0.30 \\ 
    \hline
\textsc{TStyler}$_{\textsc{recon}}$ & 0.86 & 0.15 & 0.31 & 0.32 & 0.93 & 0.15 & 0.44 & 0.37 & 0.90 & 0.13 & 0.31 & 0.28 \\ 
\textsc{TStyler}$_{\textsc{recon},\textsc{rerank}(5)}$  & 0.85 & 0.15 & 0.46 & 0.38 & 0.93 & 0.15 & 0.59 & 0.43 & 0.88 & 0.13 & 0.46 & 0.35 \\ 
\textsc{TStyler} & 0.83 & 0.13 & 0.59 & \underline{0.40} & 0.91 & 0.13 & 0.71 & \underline{0.45} & 0.84 & 0.11 & 0.58 & 0.36 \\ 
\textsc{TStyler}$_{\textsc{rerank}(5)}$ & 0.84 & 0.12 & 0.70 & \textbf{0.43} & 0.91 & 0.13 & 0.79 & \textbf{0.48} & 0.83 & 0.11 & 0.69 & \textbf{0.39} \\ 
    
    \bottomrule[\heavyrulewidth]
    \end{tabular}
    \caption{We reproduce the low-resource authorship style transfer evaluations from \citet{patel2022lowresource} on samples from the Reddit Million User Dataset \cite{khan2021deep}. We measure \textit{Away} and \textit{Towards} metrics using \textit{UAR} \cite{rivera-soto-etal-2021-learning}. We compute \textit{Sim} with MIS \cite{babakov-etal-2022-large}. The highest \textit{Joint} metrics are \textbf{bolded}. 
    }
    \label{table:uar-embedding-results}
    \end{table*}
}

\begin{figure}[t]
\includegraphics[width=7.6cm]{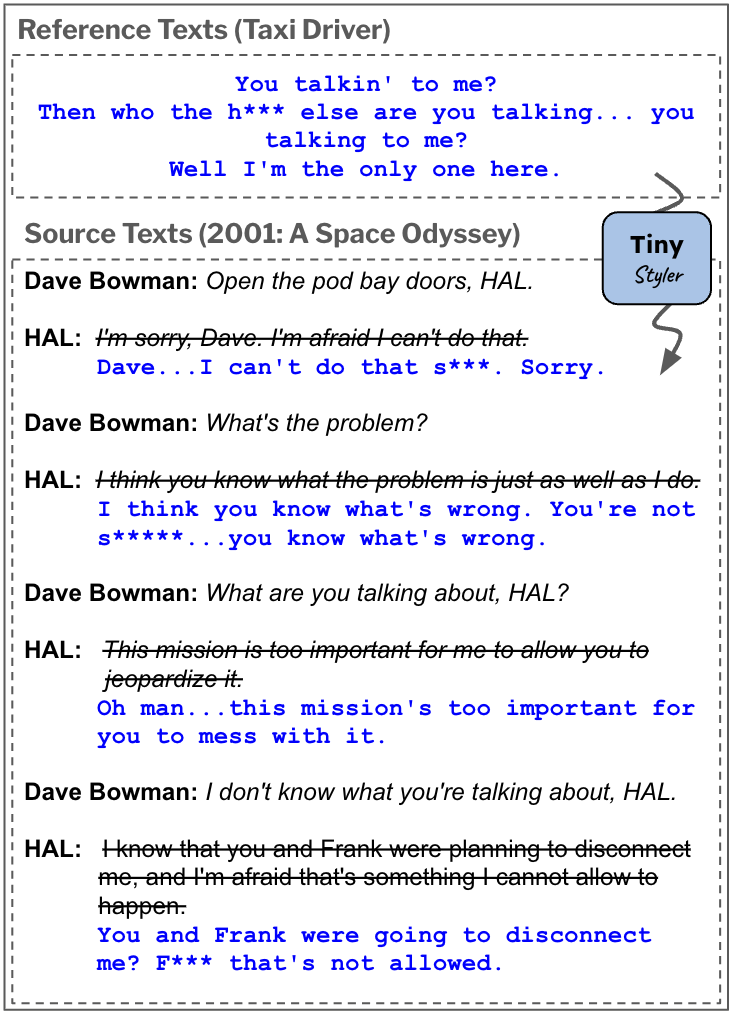}
\centering
\caption{\textsc{TinyStyler} uses authorship embeddings from examples of the target style and conditions on these to rewrite source texts to match the target style. We replace expletives above with `$\ast$'.}
\label{fig:figure1}
\end{figure}

\section{Introduction}

Text style transfer is the task of transforming a source text to match a target style while preserving its original meaning 
\cite{jin-etal-2022-deep, krishna2020reformulating, patel2022lowresource, horvitz2024paraguide}. These target styles can be defined in multiple ways, including around attributes (e.g. formality) or authorship (e.g. Barack Obama) \citep{jin-etal-2022-deep}. Building style transfer systems is complicated by the lack of paired data between different styles \citep{krishna2020reformulating}. For tasks like authorship transfer, there may even be limited available data in a target style (e.g. for a non-famous author), which poses an additional challenge \cite{patel2022lowresource} and motivates few-shot approaches.

Several recent style transfer approaches rely on prompting large language models (LLMs) \cite{patel2022lowresource,reif-etal-2022-recipe}. Unlike previous style transfer approaches, these LLM-based methods can perform well on arbitrary target styles, with only few examples of a target style. Style transfer utilizing LLMs depends on in-context learning (ICL) capabilities that only reliably emerge with scale in very large models \cite{Radford2019LanguageMA, wei2022emergent, lu2023emergent}. The inefficiency of using these large models along with long prompts packed with in-context examples limits the practical utility of these approaches. While \citet{suzgun-etal-2022-prompt} demonstrate that inference-time ranking can improve the style transfer performance of smaller language models, they also show that large performance gaps remain, particularly for rarer target styles.

Recent controllable text generation approaches present alternatives that rely on smaller models for fine-grained control over stylistic features \cite{khan2024learning, horvitz2024paraguide, mireshghallah2022mix}, however, these methods rely on slow sampling procedures or are prone to disfluencies. Moreover, these approaches are significantly more complex and cumbersome to use than prompting LLMs.

In this paper, we introduce \textsc{TinyStyler},\footnote{Our code is available at \url{https://github.com/zacharyhorvitz/TinyStyler}.} a simple and efficient approach to few-shot text style transfer that harnesses small language models and recent advances on authorship representations \cite{wegmann-etal-2022-author, rivera-soto-etal-2021-learning} that aim to capture the writing style of an author. \textsc{TinyStyler} is trained in an unsupervised fashion over a large, diverse corpus of texts to reconstruct texts from paraphrases by conditioning on an authorship embedding of the original text. At inference time, few-shot style transfer can be performed by conditioning on the authorship embedding of a new, desired target style. Inspired by RAFT \cite{dong2023raftrewardrankedfinetuning}, we further improve model performance by sampling a large number of transferred texts, filtering these texts using automatic metrics, and fine-tuning on the resulting high-quality pairs. The resulting approach enables simple, few-shot style transfer that is both on par with state-of-the-art LLMs \textit{and} fine-grained control through interpolation of the target style embedding \cite{dong2023raftrewardrankedfinetuning}.

\textbf{In summary, our contributions are as follows:}

\begin{figure*}[t]
\includegraphics[width=\textwidth]{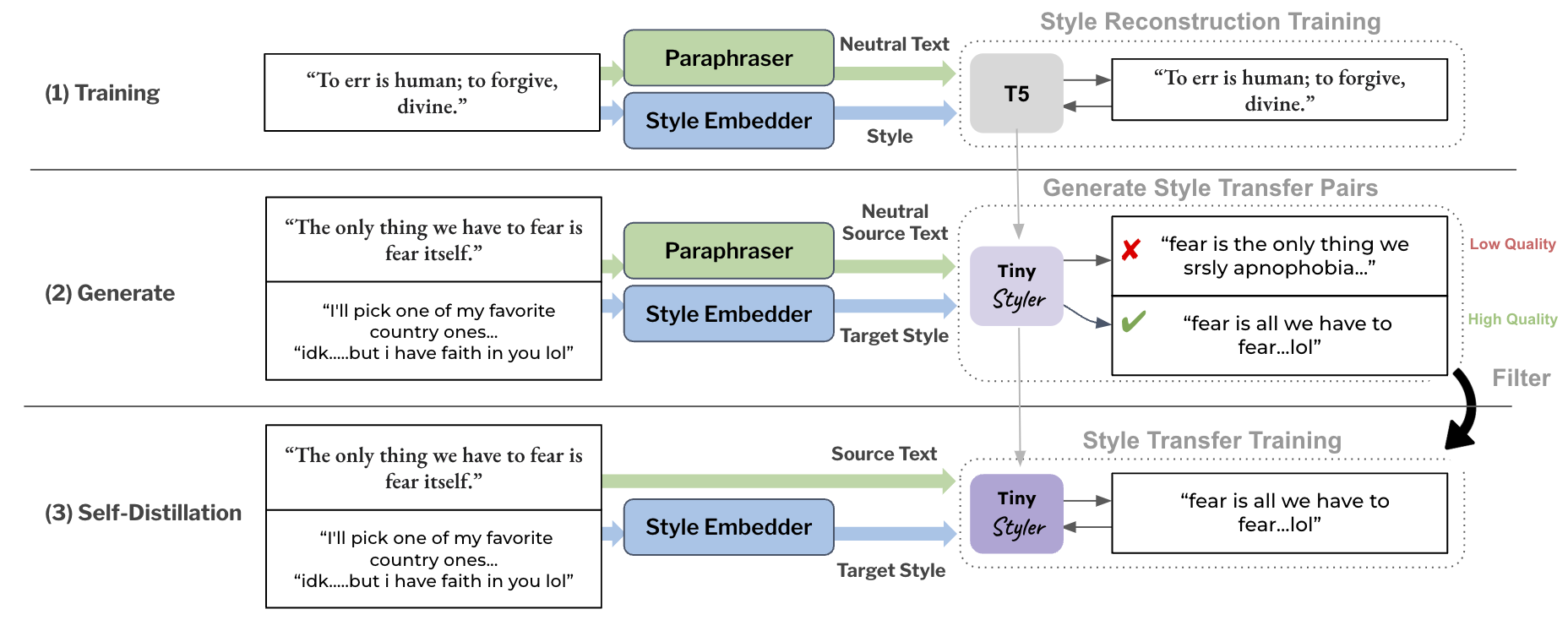}
\centering
\caption{\textbf{Step 1)} We train a model to reconstruct texts from their paraphrases following \citet{krishna2020reformulating}, however, we only train a single model for all styles. To do this, we condition reconstruction on pre-trained authorship embeddings. \textbf{Step 2)} We generate style transfer pairs by transforming Reddit posts from a source author by conditioning generation on authorship embeddings from a different Reddit author. We filter low-quality style transfer pairs automatically using meaning preservation and stylistic similarity metrics. \textbf{Step 3)} We self-distill our model on the remaining high-quality pairs to improve the consistency of our approach and remove the reliance on a separate, external paraphrasing model.}
\label{fig:figure2}
\end{figure*}

\begin{enumerate}
    \item We introduce \textsc{TinyStyler}, a fast, efficient, and performant approach for few-shot style transfer with authorship embeddings.

        \item We evaluate our approach on both authorship and formality style transfer tasks, where we find that \textsc{TinyStyler} outperforms or is competitive with strong baselines like \textsc{Gpt-3.5} and \textsc{Gpt-4}, as well as outperforms recent controllable style transfer methods. Our human evaluation provides evidence that \textsc{TinyStyler} (with only 800M parameters) offers an efficient alternative to  state-of-the-art LLMs and in-context learning.

         \item  By conditioning on an interpolation between the source style embedding and the target style embedding, we find \textsc{TinyStyler} enables fine-grained control over trading-off style transfer accuracy for meaning preservation. As a result, the approach confers many of the benefits of alternative recent controllable generation approaches \cite{khan2024learning, horvitz2024paraguide, mireshghallah2022mix}.

\end{enumerate}

\section{Related Work}

\paragraph{Unsupervised Style Transfer} Reconstructing text as a framework for performing unsupervised style transfer where no parallel examples between the source and target styles exist is an established pattern used in prior work \cite{krishna2020reformulating, riley-etal-2021-textsettr, jangra-etal-2022-star, horvitz2024paraguide}. In this paper, we build on the framework introduced by \citet{krishna2020reformulating}, which first uses paraphrasing to neutralize the style of a text and then trains a reconstruction model that learns to re-stylize it. Rather than train a model per target style, we perform style transfer to different target styles with a single model by conditioning on a representation of the target style like \citet{riley-etal-2021-textsettr}. Unlike both approaches, we leverage the information captured in strong, pre-trained authorship embeddings \cite{wegmann-etal-2022-author, rivera-soto-etal-2021-learning}.

\paragraph{Few-Shot Style Transfer with LLMs} Several recent approaches perform style transfer with LLMs and in-context learning \cite{patel2022lowresource, reif-etal-2022-recipe}. \citet{suzgun-etal-2022-prompt} investigates reranking as a method to boost the quality of outputs generated by smaller LLMs.  Other methods perform knowledge distillation from larger LLMs into smaller models \cite{saakyan2023iclef, zhang2024distilling}.

\paragraph{Controllable Text Generation} Another line of recent work has applied controllable text generation approaches to style transfer \cite{horvitz2024paraguide,khan2024learning, kumar2021controlled,mireshghallah2022mix, dale2021text}. 
While text diffusion approaches like \textsc{ParaGuide} \cite{horvitz2024paraguide} and MCMC approaches like Mix \& Match \cite{mireshghallah2022mix} afford fine-grained stylistic control, their non-autoregressive sampling procedures corresponds to longer inference times and an increased risk of disfluent outputs.

\section{\textsc{TinyStyler}}

Our style-transfer approach is built around reconstructing texts. Following previous work \cite{krishna2020reformulating,patel2022lowresource, horvitz2024paraguide}, we use paraphrasing to neutralize stylistic features from texts while preserving their original meaning. In Section \ref{sec:stylereconstruction}, we describe training a single model to reconstruct these texts from both their paraphrases \textit{and} their pre-trained authorship embeddings. In Section \ref{sec:generatingstyletransferpairs}, we describe how we can then transform source texts to arbitrary target styles with the trained reconstruction model by conditioning on new authorship embeddings from texts in a target style. We use this reconstruction model to construct a dataset of high-quality synthetic style transfer pairs. Finally, in Section \ref{sec:selfdistillation}, we perform self-distillation to improve the consistency of our model and remove the reliance on a separate paraphrasing model. We illustrate the \textsc{TinyStyler} procedure in Figure \ref{fig:figure2}.

\subsection{Style Reconstruction Training}
\label{sec:stylereconstruction}

Following \citet{krishna2020reformulating}, we first train a model to reconstruct a text from a paraphrase of the original text with neutralized style. Instead of training a unique model per target style, we train a single reconstruction model that conditions on both a paraphrase and an authorship embedding for a source text  (See Figure \ref{fig:figure2}, Step 1). Later, in \ref{sec:generatingstyletransferpairs}, we leverage this approach to build a style transfer dataset for training a stronger, simpler pipeline.

\paragraph{Dataset} Training a general-purpose reconstruction model requires a corpus that covers many diverse authorship styles. Accordingly, we use a subset of the Reddit Million User Dataset (MUD) \cite{khan2021deep},
which contains comments from over $1$ million Reddit users. For each username, we sample $10$ random comments and filter all comments longer than $60$ tokens. The resulting dataset contains $8$ million comments in total.

\paragraph{Authorship Embeddings} %
We consider two approaches that learn neural representations of the writing style of authors in continuous space:
1) ``\textsc{Style} Embeddings'' \cite{wegmann-etal-2022-author} and 2) ``Universal Authorship Representations (\textsc{UAR})'' \cite{rivera-soto-etal-2021-learning}. 
A critical property of representations for style transfer is that they disentangle style and content. \textsc{Style} embeddings, for example, are trained with a contrastive authorship verification (CAV) objective, using texts from the same author on different topics as positive examples and texts on the same topic from different authors as negative examples to form training triplets.  As a result, we train \textsc{TinyStyler} to perform style transfer by conditioning on \textsc{Style} embeddings. We reserve \textsc{UAR} embeddings, which are trained on a larger dataset, as a held-out authorship representation space for automatic evaluation of authorship style transfer following  \citet{patel2022lowresource} and \citet{horvitz2024paraguide}.

\begin{table*}[h]
\centering
\renewcommand{\arraystretch}{1.5} %
\fontsize{8}{10}\selectfont
\begin{tabular}{>{\raggedright\arraybackslash}p{3cm}>{\raggedright\arraybackslash}p{3cm}>{\raggedright\arraybackslash}p{3cm}>{\raggedright\arraybackslash}p{3cm}}
\toprule
\textbf{Source} & \textbf{Informal} & \textbf{Question + Sentence} & \textbf{Barack Obama's Speeches} \\ \midrule
\textit{\textbf{Toto, I've a feeling we're not in Kansas anymore.}} & \textit{i think we arent in Kansas anymore tbh.} & \textit{We are not in Kansas anymore? I feel you.} & \textit{I think we are no longer in Kansas.} \\ \midrule
\textit{\textbf{is mayonnaise an instrument?}} & \textit{oh wait mayonnaise is an instrument :(} & \textit{Can you tell me what mayonnaise is? It is an instrument.} & \textit{This makes me wonder if mayonnaise is an instrument.} \\ \midrule
\textit{\textbf{Life is like riding a bicycle. To keep your balance, you must keep moving.}} & \textit{life is like riding a bicycle so you gotta keep moving :(} & \textit{Life is like riding a bicycle. Do you have to keep moving?} & \textit{But life is like riding a bicycle - you must keep moving.} \\ \midrule
\textit{\textbf{Life moves pretty fast. If you don't stop and look around once in a while, you could miss it.}} & \textit{life moves fast if you dont stop and look around once in a while i think.} & \textit{Is this an important lesson? Life moves fast. If you don't stop and look around once in a while, you could miss it.} & \textit{If you don't stop and look around once in a while, you could miss it.} \\ \midrule
\textit{\textbf{The first rule of Fight Club is you do not talk about Fight Club.}} & \textit{u dont talk about fight club you know,first rule.} & \textit{First rule of Fight Club? Do not talk about Fight Club.} & \textit{The first rule of Fight Club is not to talk about it.} \\ 
\bottomrule
\end{tabular}
\caption{We display \textsc{TinyStyler} outputs for various target styles. These outputs demonstrate \textsc{TinyStyler}'s ability to transform text across various stylistic properties from lexical and punctuation choice to syntactic structure.}
\label{tab:quotes}
\end{table*}

\vspace{-4pt}
\paragraph{Architecture} To reconstruct texts from paraphrases and authorship embeddings, we fine-tune a modified \textsc{T5} model \cite{raffel2020exploring} with 800 million parameters. We adapt the model to condition on authorship style information in the authorship embedding by jointly learning a projection from the embedding's dimension ($d=768$) to the \textsc{T5} model's hidden dimension ($d=512$). We then prepend the projected embedding to the word embeddings of the input text.

\vspace{-3pt}
\paragraph{Training Details}
We generate paraphrases and \textsc{Style} embeddings for each comment in our corpus. To generate paraphrases, we use an off-the-shelf paraphrasing \textsc{Pegasus} model \cite{zhang2020pegasus} with the same configuration as \citet{horvitz2024paraguide}.\footnote{We use \texttt{\small tuner007/pegasus\_paraphrase}.} We train the T5 model to reconstruct the original comment, conditioned on each (paraphrase, \textsc{Style} embedding) pair.  We include additional details on our training procedure and hyperparameters in Appendix \ref{sec:style-recon-training}.

\uarembeddingresultstable

\subsection{Generating Style Transfer Pairs}
\label{sec:generatingstyletransferpairs}

The reconstruction model can now be used for style transfer by paraphrasing a text and using the model for reconstruction while conditioning on the \textsc{Style} embedding of a desired target style (See Figure \ref{fig:figure2}, Step 2). To use multiple example texts of a target style, we combine their \textsc{Style}  embeddings through a mean pool operation. This enables our approach to condition on an arbitrary number of example texts of a target style with no additional memory overhead. This initial approach to style transfer is already fast and efficient compared to text diffusion denoising \citep{horvitz2024paraguide} or MCMC sampling \citep{mireshghallah2022mix}, and inexpensive compared with prompting LLMs. We take advantage of this reconstruction model's efficiency at performing style transfer and generate many example style transfer pairs, rerank them using automatic evaluation metrics for quality, and filter out low-quality pairs. The result is a synthetic dataset of high-quality examples of style transfer.

\paragraph{High Quality Dataset} To build our high quality synthetic dataset, we sample $160$k unique random author pairs from the Reddit MUD Dataset \cite{khan2021deep}. For each pair, we choose a random source text and generate multiple outputs by sampling different paraphrases from the paraphrase model. \citet{jangra-etal-2022-star} found that models trained to reconstruct from paraphrases are prone to hallucinations and we hypothesize sampling different paraphrases and reranking model outputs may mitigate these hallucinations.

\paragraph{Reranking and Filtering}

Like \citet{suzgun-etal-2022-prompt}, we rank outputs from our style transfer system using automatic metrics. To rank outputs, we utilize the automatic style transfer metrics (\textit{Away}, \textit{Towards}, and \textit{Sim}) introduced by \citet{patel2022lowresource}. We rank each output per inference using the geometric mean of all three metrics ($G(G(Away,Towards), Sim)$) and select the output with the highest score. We compute \textit{Away} and \textit{Towards} scores using \textsc{Style} embeddings \cite{wegmann-etal-2022-author}. To compute \textit{Sim} scores, we use Mutual Implication Score (MIS), which has been shown to correlate with human judgments on style transfer tasks \cite{babakov-etal-2022-large}. We then filter the resulting outputs with low scores on two meaning preservation metrics, MIS \cite{babakov-etal-2022-large} and SimCSE \cite{gao2022simcse}. We also filter outputs with low \textit{Away} and \textit{Towards} metrics, which we compute with \textsc{Style} embeddings \cite{wegmann-etal-2022-author}. When multiple candidates remain, we select the highest ranked example. After filtering and selection, we are left with $40$K high-quality examples of style transfer pairs. Additional details on dataset generation are included in Appendix \ref{sec:dataset-generation}.

\subsection{Self-Distillation on High Quality Examples}
\label{sec:selfdistillation}

The style transfer procedure we describe in Section \ref{sec:generatingstyletransferpairs} with the trained reconstruction model still relies on using a paraphrase model to generate inputs for the reconstruction model. This requires performing inference with a separate model, a cumbersome procedure that also increases inference times. Additionally, while we can utilize reranking to improve performance, generating multiple candidate outputs also requires longer inference times and additional compute overhead \citep{suzgun-etal-2022-prompt}. To address these limitations and improve the consistency of our approach, we distill away reranking and the paraphrasing step entirely by further fine-tuning our reconstruction model on the high-quality synthetic dataset generated by reconstruction model itself, essentially a self-distillation \cite{DBLP:journals/corr/abs-1905-08094} (See Figure \ref{fig:figure2}, Step 3). During self-distillation, we fine-tune the reconstruction model to generate the high-quality output found through reranking and filtering and condition generation on the source text and the target author's \textsc{Style} embeddings. Additional details on our self-distillation procedure are in Appendix \ref{sec:self-distill-ft}.

\begin{table*}[t]
\fontsize{8}{10}\selectfont
\centering

\begin{tabular}{lcccccc}
\toprule
Method & Acc ($\rightarrow \textit{F}$,$\rightarrow \textit{I}$) &  Sim ($\rightarrow \textit{F}$,$\rightarrow \textit{I}$) & Fluency ($\rightarrow \textit{F}$,$\rightarrow \textit{I}$) & Joint ($\rightarrow \textit{F}$,$\rightarrow \textit{I}$) & GPT-2  \\
\midrule
\textsc{Copy}$_{\textsc{src}}$ & 0.06 (0.10, 0.01) & 0.96 (0.96, 0.97) & 0.80 (0.71, 0.88) & 0.05 (0.09, 0.01) & 97.14  \\
\midrule
\multicolumn{6}{l}{\textit{Large Language Models}} \\
\midrule
\textsc{Gpt-3.5} & 0.90 (0.97, 0.82) & 0.86 (0.86, 0.87) & 0.85 (0.91, 0.79) & 0.79 (0.89, 0.69) & 76.53  \\
\textsc{Gpt-4} & 0.95 (0.99, 0.91) & 0.89 (0.87, 0.91) & 0.84 (0.91, 0.78) & 0.85 (0.90, 0.80) & 101.43 \\
\midrule
\multicolumn{6}{l}{\textit{Controllable Text Generation Methods}} \\
\cmidrule{1-6}
$\textsc{M\&M}_{\textsc{Disc}}$ & 0.52 (0.12, 0.92) & 0.38 (0.37, 0.38) & 0.52 (0.52, 0.53) & 0.24 (0.06, 0.43) & 167.18 \\
$\textsc{M\&M}_{\textsc{Ham}}$ & 0.49 (0.08, 0.90) & 0.56 (0.56, 0.57) & 0.50 (0.48, 0.52) & 0.29 (0.05, 0.53) & 191.08  \\
$\textsc{PGuide}_{\lambda=200}$ & 0.94 (0.91, 0.96) & 0.65 (0.61, 0.69) & 0.69 (0.68, 0.70) & 0.68 (0.64, 0.71) & 160.15 \\
$\textsc{PGuide}_{\lambda=1000}$ & 0.97 (0.95, 0.99) & 0.56 (0.47, 0.65) & 0.60 (0.58, 0.63) & 0.61 (0.54, 0.68) & 280.54  \\
$\textsc{PGuide}_{\lambda=5000}$  & \textbf{0.97} (0.95, 0.99) & 0.49 (0.37, 0.61) & 0.54 (0.51, 0.57) & 0.55 (0.46, 0.64) & 503.46 \\
\midrule
$\textsc{TStyler}$ & 0.92 (0.88, 0.97) & 0.80 (0.80, 0.81) & 0.77 (0.82, 0.72) & 0.76 (0.74, 0.79) & \textbf{111.58} \\
$\textsc{TStyler}_{\textsc{ex}=64}$ & 0.94 (0.90, 0.98) & \textbf{0.82} (0.81, 0.82) & \textbf{0.77} (0.83, 0.72) & \textbf{0.78} (0.77, 0.80) & 112.5  \\
\bottomrule
\end{tabular}

\caption{We evaluate formality style transfer on GYAFC and perform an automatic evaluation. We separate performance toward a formal target style ($\rightarrow F$) and informal target style ($\rightarrow I$). The best controllable approach result for each metric is \textbf{bolded}.
\\
}
\label{table:formal_auto}
\end{table*}

\begin{table*}[t]
\fontsize{8}{10}\selectfont
\centering
\begin{tabular}{lcccccc}
\toprule
Method & Acc ($\rightarrow \textit{F}$,$\rightarrow \textit{I}$) &  Sim ($\rightarrow \textit{F}$,$\rightarrow \textit{I}$) & Fluency ($\rightarrow \textit{F}$,$\rightarrow \textit{I}$) & Joint ($\rightarrow \textit{F}$,$\rightarrow \textit{I}$) & \\
\midrule

\textsc{Gpt-3.5} & 0.94 (0.96, 0.92) & 0.97 (0.96, 0.97) & 1.00 (1.00, 1.00) & 0.91 (0.92, 0.89) \\
\textsc{Gpt-4} & 0.95 (1.00, 0.89) & 0.97 (0.95, 0.99) & 0.99 (1.00, 0.99) & 0.91 (0.95, 0.88) \\
\midrule
$\textsc{M\&M}_{\textsc{Ham}}$ & 0.48 (0.05, 0.91) & 0.26 (0.21, 0.31) & 0.42 (0.35, 0.49) & 0.11 (0.01, 0.21) \\
$\textsc{PGuide}_{\lambda=200}$ & 0.88 (0.89, 0.87) & 0.48 (0.49, 0.47) & 0.84 (0.85, 0.83) & 0.39 (0.43, 0.36) \\
$\textsc{TStyler}_{\textsc{ex}=64}$ & \textbf{0.89} (0.80, 0.97) & \textbf{0.77} (0.67, 0.87) & \textbf{0.98} (0.96, 1.00) & \textbf{0.71} (0.56, 0.85) \\

\bottomrule
\end{tabular}
\caption{Human annotator ratings of formality style transfer outputs over GYAFC on formality (\textit{Accuracy}), meaning preservation (\textit{Similarity}), and \textit{Fluency}. \textit{Joint} averages the metrics on a per-example basis.}
\label{table:formal_human}
\end{table*}

\section{Evaluation}

We evaluate \textsc{TinyStyler} on authorship style transfer and text attribute style transfer (formal $\leftrightarrow$ informal).

\subsection{Low-Resource Authorship Transfer}

For non-famous authors, there may only be a few texts available in their authorship style. Low-resource authorship style transfer is the task of transforming to a target author's style with limited target style data. We evaluate our approach on the dataset introduced by \citet{patel2022lowresource} using their three dataset splits (\textit{Random}, \textit{Single}, \textit{Diverse}). Each split has $15$ Reddit users that serve as source styles, $15$ Reddit users that serve as target styles, each with $16$ writing samples. In total, there are 225 (15 * 15) style transfer directions and 3600 (225 * 16) total transformations per split.

\vspace{-4pt}
\paragraph{Implementation Details} 
To perform authorship style transfer with \textsc{TinyStyler}, we condition on the target style examples using a mean pool operation over the embeddings. We consider two configurations of \textsc{TinyStyler}. \textsc{TinyStyler}$_\textsc{recon}$ is the initial style transfer approach detailed in Section \ref{sec:generatingstyletransferpairs} and Figure \ref{fig:figure2}, Step 2 that reconstructs from a paraphrase and target style embedding. \textsc{TinyStyler} is the final model secondarily fine-tuned on the high quality synthetic dataset described in Section \ref{sec:selfdistillation} to distill away the use of the separate paraphraser model and improve consistency and quality. For each configuration, we also investigate the effect of reranking at inference time as an optional technique to further boost performance.
While inference time reranking adds additional compute overhead, we note that our approach has lower inference times than other techniques (See Appendix \ref{sec:timing}). Appendix \ref{sec:authorship-eval} discusses the authorship transfer evaluations in detail.

\vspace{-3pt}

\paragraph{Metrics} 
We evaluate on the \textit{Away}, \textit{Towards}, \textit{Sim}, and \textit{Joint} metrics from \citet{patel2022lowresource}.  Notably, unlike during reranking, where we compute these same metrics like \textit{Away} and \textit{Towards} with \textsc{Style} embeddings, during evaluation we instead compute these metrics using the held-out
\textsc{UAR} embeddings \cite{rivera-soto-etal-2021-learning}. This avoids directly reranking on the automatic style evaluation metric, which could inflate performance.

\paragraph{Baselines} 
We include results for all methods implemented by \citet{patel2022lowresource}, including LLM-based approaches like \textsc{Styll}$_{\textsc{Gpt-3}}$ and \textsc{Styll}$_{\textsc{Bloom}}$. Additionally, we include results for \textsc{ParaGuide}, a recent style transfer approach that uses text diffusion models \cite{horvitz2024paraguide}. We also include results from prompted \textsc{Gpt-3.5} and \textsc{Gpt-4}. Additional details on our baseline implementations are included in Appendix \ref{sec:baselines}.

\subsection{Formality Transfer}

We also evaluate \textsc{TinyStyler}'s ability to perform text attribute style transfer using the established GYAFC dataset \cite{rao-tetreault-2018-dear}. Authorship style transfer is difficult for humans to evaluate \citep{patel2022lowresource}, and we select formality as an additional task because the attribute is broadly recognizable to human annotators.

\paragraph{Implementation Details}

We perform formality style transfer with \textsc{TinyStyler} by providing few-shot examples of formal or informal texts. We condition on $\textsc{num examples}$ of texts by extracting style embeddings for each these target style examples and then mean pooling their embeddings. We evaluate \textsc{TinyStyler} with \textsc{num examples} $=16$ and \textsc{num examples} $=64$. 
Several of the baselines that we compare against require a classifier trained on formal and informal texts to guide generation. We train a model for this purpose on GYAFC, and also use this classifier to select representative few-shot examples for \textsc{TinyStyler}.

\paragraph{Baselines} We compare \textsc{TinyStyler} to \textsc{ParaGuide} \cite{horvitz2024paraguide} and \textsc{Mix and Match} \cite{mireshghallah2022mix}, which are two recent controllable generation approaches. Both approaches are guided at inference time by the classifier trained on GYAFC. We also benchmark against \textsc{Gpt-4} and \textsc{Gpt-3.5}, as well as a naive \textsc{COPY} baseline for reference.

\paragraph{Metrics}
To evaluate style transfer accuracy, we use an off-the-shelf formality classifier that is held-out for evaluation \cite{dementieva-etal-2023-detecting,briakou2021xformal}. Similar to our authorship style transfer evaluations, we measure meaning preservation (\textit{Sim}) using Mutual Implication Score (MIS) \cite{babakov-etal-2022-large}. Because many of the controllable approaches baselines are prone to disfluencies \cite{horvitz2024paraguide, mireshghallah2022mix}, we also compute \textit{Fluency} scores using a model trained on the CoLA Dataset \cite{morris2020textattack,warstadt-etal-2019-neural}. We also report median \textit{GPT-2 Perplexity} metrics which has been proposed as an alternative fluency metric \citep{khan2024learning}. To compute an aggregate \textit{Joint} metric, we follow \citet{horvitz2024paraguide} and \citet{krishna2020reformulating}, and compute the geometric mean of \textit{Accuracy}, \textit{Sim}, and \textit{Fluency}. 

\paragraph{Human Evaluation} To validate that our evaluation sufficiently aligns with human judgment of style transfer quality, we ask multiple annotators to score outputs from our approach against outputs from various approaches. We asked annotators to evaluate meaning preservation, fluency, and the formality of model outputs with binary judgements. More details describing our human evaluations are included in Appendix \ref{sec:human_eval}.

\section{Results}

\subsection{Authorship Transfer}

Table \ref{table:uar-embedding-results} contains our authorship transfer results. Even without reranking or supervised distillation, \textsc{TinyStyler}$_{\textsc{recon}}$ is competitive with the other approaches. This unsupervised approach outperforms all non-LLM baselines on \textit{Joint} metrics.  The method has strong \textit{Away} and \textit{Towards} metrics, but comparably low meaning preservation  (\textit{Sim}) scores. These low \textit{Sim} scores are addressed by our refined approach, \textsc{TinyStyler}. The self-distilled \textsc{TinyStyler} achieves much higher \textit{Joint} scores, largely due to improvements on its meaning preservation, which is comparable to LLM-based approaches. \textsc{TinyStyler} outperforms almost all baselines on \textit{Joint} metrics. The only exception is that it slightly under-performs \textsc{Gpt-3.5} on the \textit{Diverse} evaluation subset. With additional inference-time reranking, \textsc{TinyStyler}$_{\textsc{rerank}(5)}$ widens this gap, and outperforms all methods on all evaluation sets.

We find \textsc{TinyStyler} demonstrates strong performance on authorship style transfer and outperforms LLMs with a notably lightweight approach, using only \textasciitilde  $0.5\%$\ the parameters of \textsc{Gpt-3.5}. 
Additionally, our experiments ablating reranking demonstrate its effectiveness in improving meaning preservation.

\begin{figure}[t]
\includegraphics[width=7cm]{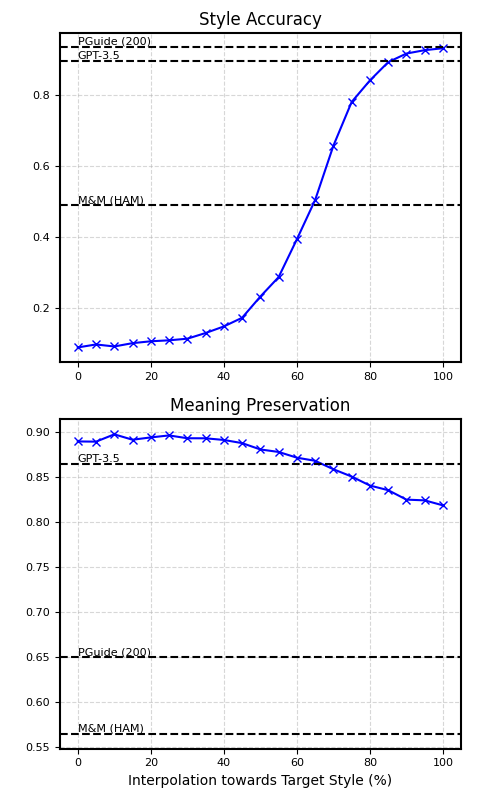}
\centering
\caption{\textsc{TinyStyler} affords control over the strength of style transfer by interpolating between the source and target styles in \textsc{Style} embedding  space. The effect on style transfer metrics for different degrees of interpolation are visualized using GYAFC.
}
\label{fig:interp}
\end{figure}

\begin{figure}[t]
\includegraphics[width=7.6cm]{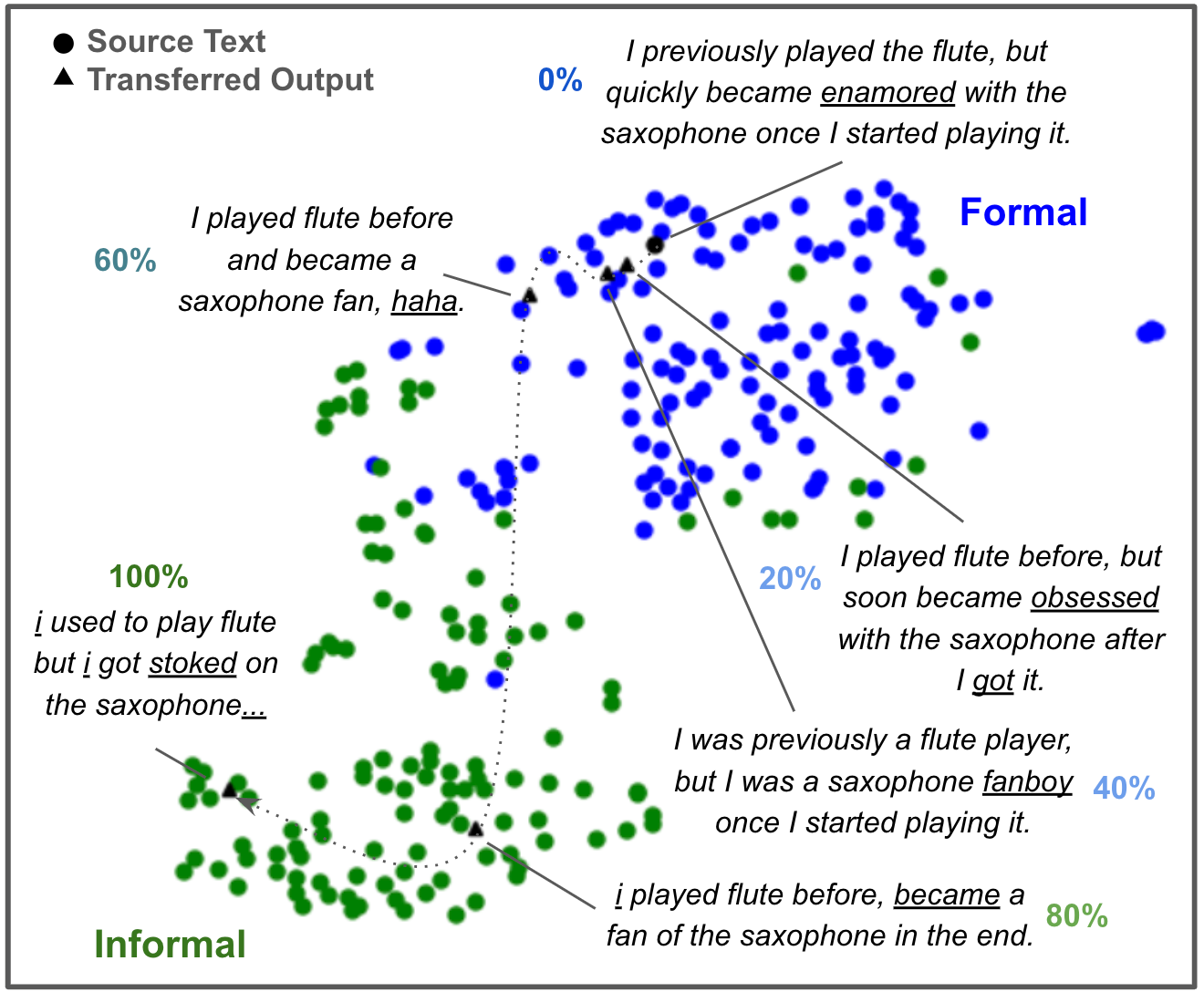}
\centering
\caption{We transform a formal text in GYAFC by interpolating (0\% to 100\%) towards the average embedding of the informal texts with \textsc{TinyStyler}. We visualize the outputs alongside samples from the corpus using a t-SNE projection \cite{JMLR:v9:vandermaaten08a}. Texts are embedded with \textsc{style} embeddings. 
}
\label{fig:tsne}
\end{figure}

\subsection{Formality Transfer}

Tables \ref{table:formal_auto} and \ref{table:formal_human} contain formality transfer results. In both automatic and human evaluations, \textsc{TinyStyler} outperforms all controllable baselines on \textit{Joint}, \textit{Fluency}, \textit{GPT-2 Perplexity} and \textit{Sim} metrics. Because \textsc{TinyStyler} aggregates target style embeddings via mean pooling, the approach can condition on additional target examples with no memory overhead. This additional conditioning in $\textsc{TStyler}_{\textsc{ex}=64}$ corresponds to further improvements on \textit{Accuracy} and \textit{Joint} metrics.  While \textsc{ParaGuide} has comparable \textit{Accuracy} scores to \textsc{TinyStyler}, the approach was rated as significantly less fluent and meaning preserving.

Considering all approaches, \textsc{Gpt-4} and \textsc{Gpt-3.5} are most performant on \textit{Joint}, \textit{Sim}, and \textit{Fluency} metrics. 
The strong performance of LLMs on this dataset is unsurprising, as these models are trained on a large portion of the internet \cite{Achiam2023GPT4TR,chatgptold}, which contains many examples of informal and formal texts. Contamination of GYAFC data or other formal re-writing examples in instruction-tuning supervised training data may also inflate performance of these models \cite{sainz-etal-2023-nlp}. Despite these disadvantages, \textsc{TinyStyler} performs competitively with these larger models on \textit{Accuracy} and \textit{Fluency} ratings. Additionally, on informal transfer, \textsc{TinyStyler} performs on par with \textsc{Gpt-4} on the \textit{Joint} metric ($0.85$ vs $0.88$).

\paragraph{\textbf{Inference Timing}} We include
inference timing results in our Appendix. \textsc{TinyStyler} offers subsecond inference times on a single A$100$ GPU, and runs >$35$x faster than all controllable baselines. The approach is >$1.5$x faster than the LLM methods, which we evaluated through APIs. These results indicate that \textsc{TinyStyler} can be readily deployed in time-constrained practical applications.

\subsection{Interpolating in Style Space}

Like other controllable text generation approaches \cite{khan2024learning, horvitz2024paraguide, mireshghallah2022mix}, \textsc{TinyStyler} enables specifying target styles at inference time. However, another advantage of these controllable methods over prompt-based style transfer is that they enable direct control of the trade-offs between style transfer and meaning preservation. In Figure \ref{fig:interp}, we visualize the effect of interpolating in \textsc{Style} embedding space style on formal $\leftrightarrow$ informal transfer metrics. These results indicate that  \textsc{TinyStyler} also affords control of the balance between metrics. Moving away from the source text embedding and towards the target style increases style transfer accuracy at the expense of meaning preservation. In Figure \ref{fig:tsne}, we visualize the effect of embedding interpolation on the output text, where movement in stylistic space corresponds to typographical, lexical, and syntactic changes.

\section{Conclusion and Future Work}

We introduce \textsc{TinyStyler}, a fast, simple to use, efficient approach to few-shot style transfer that uses small language models and pre-trained authorship embeddings. The method outperforms strong baselines, including \textsc{Gpt-4}, on authorship transfer. The method also outperforms other controllable approaches on formality transfer, and is more competitive with LLMs. Our work highlights the utility of pre-trained standalone authorship embeddings and we look forward to future work on representations that capture more diverse characteristics of text style. Additionally, \textsc{TinyStyler} showcases the potential value of reranking with automatic evaluation metrics, filtering, and self-distillation as a procedure towards training efficient models that can compete with and close the performance gap with LLMs. Accordingly, we are enthusiastic about the potential for future improvements to automatic text evaluation yielding performance benefits when incorporated into text style transfer and other text generation pipelines.

\section{Limitations}

\textsc{TinyStyler} leverages pre-trained authorship embeddings. While \textsc{TinyStyler} can benefit from continued advances in authorship style representation learning, the approach is also bottlenecked by their current representational capacity. As a result, \textsc{TinyStyler} may underperform on more rare stylistic choices (e.g. iambic pentameter) that are not captured by \textsc{Style} embeddings. Additionally, authorship style transfer metrics may not fully capture the preferences of human authors.

\section{Ethical Considerations}
\textsc{TinyStyler} is an efficient approach for few-shot style transfer that consumes far fewer resources than alternative LLM-based methods. Consequently, the method can empower individuals and organizations to rewrite texts or personalize generic outputs from chat models to match a user's preferences. Simultaneously, efficient text style transfer can aid malicious actors with impersonation. Accordingly, research on text style transfer warrants a renewed focus on AI generated text detection and investment in the media literacy of the broader public.

\section{Acknowledgements}

We would like to thank Rahul Aditya, Amith Ananthram, Debasmita Bhattacharya, Yanda Chen, Nicholas Deas, Fei-Tzin Lee, Melanie Subbiah, Elsbeth Turcan, Haoda Wang
and Yunfan Zhang for help with our human evaluations. We would also like to thank our anonymous reviewers for their constructive feedback. This research is supported in part by the Office of the Director of National Intelligence (ODNI), Intelligence Advanced Research Projects Activity (IARPA), via the HIATUS Program contract \#2022-22072200005. The views and conclusions contained herein are those of the authors and should not be interpreted as necessarily representing the official policies, either expressed or implied, of ODNI, IARPA, or the U.S. Government. The U.S. Government is authorized to reproduce and distribute reprints for governmental purposes notwithstanding any copyright annotation therein.

\bibliography{custom}

\appendix

\section{\textsc{TinyStyler} Details}
\label{sec:appendix}

\subsection{Style Reconstruction Training}
\label{sec:style-recon-training}

These sections provide details on training \textsc{TinyStyler}$_\textsc{recon}$ to reconstruct texts from paraphrases and \textsc{Style} embeddings.

\subsubsection{Dataset}

We construct our training and validation datasets from the publicly available Reddit Million User Dataset \cite{khan2021deep}. We reduce the size of the dataset by randomly sampling $10$ comments per user. We then filter all comments longer than $60$ tokens using the \textsc{Pegasus} tokenizer \cite{zhang2020pegasus}, resulting in $8$ million texts. We divide users into train/validation/test ($0.90,0.05,0.05$), and ensure that no users from our evaluation datasets are included in the set of training authors.

\subsubsection{Paraphrase and \textsc{Style} Embedding Generation}
\label{sec:para-generation}

We generate paraphrases with a popular off-the-shelf \textsc{Pegasus} model \cite{zhang2020pegasus} that was fine-tuned for paraphrasing.\footnote{\url{https://huggingface.co/tuner007/pegasus_paraphrase}} We use the same paraphrase model and inference hyperparameters as in \textsc{ParaGuide} \cite{horvitz2024paraguide}. To paraphrase each text in our training corpus, we sample from the paraphrase model by perfoming nucleus sampling \cite{holtzman2020curious} with $\texttt{top-p}=0.80$ and $\tau=1.5$, on a beam search of size $8$. We extract \textsc{Style} embeddings \cite{wegmann-etal-2022-author} for each comment in the $8$ million sample using the publicly available checkpoint and inference logic.\footnote{\url{https://huggingface.co/AnnaWegmann/Style-Embedding}}

\subsubsection{Architecture}

We modify a T5-Large \cite{raffel2020exploring}\footnote{\url{https://huggingface.co/google/t5-v1_1-large}}($800$ million parameters) to condition on an input text and \textsc{Style} embedding. To incorporate the \textsc{Style} embedding, we project the vector from $d=768$ to the model's embedding size ($d=412$), and preprend the result to the input word embeddings.

\subsubsection{Training Hyperparameters}

We fine-tune the modified T5 \cite{raffel2020exploring} model with the hyperparameters in Table \ref{table:hparam} on an NVIDIA-A100 GPU. We performed minimal hyperparameter tuning, instead using established learning rates and batch sizes from previous work \cite{horvitz2024paraguide}.
\begin{table}[h!]
\centering
\begin{tabular}{@{}lc@{}}
\toprule
\textbf{Hyperparameter} & \textbf{Value} \\ \midrule
Pretrained Ckpt & \texttt{\small google/t5-v1\_1-large} \\
Learning Rate & $1 \times 10^{-5}$ \\
Batch Size & 16 \\
Grad Accum. & 4 \\
Optimizer & Adam \\
Weight Decay & 0.01 \\
Schedule & Constant \\
Warm-up Steps & 2000 \\
Total Steps & 230,000 \\ \bottomrule
\end{tabular}
\caption{Fine-tuning hyperparameters for \textsc{TinyStyler}$_{\textsc{recon}}$.}
\label{table:hparam}
\end{table}
This model is trained to reconstruct the original comment from its paraphrase and \textsc{Style} embedding. We jointly learn the \textsc{Style} embedding projection alongside the other model parameters.

\subsection{Self-Distillation Details}

The following sections describe the data generation and fine-tuning procedure for self-distillation.

\subsubsection{Data Generation with \textsc{TinyStyler}$_\textsc{recon}$}
\label{sec:dataset-generation}

To build our \textbf{High Quality Dataset}, we follow the following procedure:

\begin{enumerate}
    \item We first randomly sample $160$k unique random (source, target) author pairs from our training dataset. 

    \item For each source author, we sample a single source text. 

    \item For each source text, we generate $5$ paraphrases (As in \ref{sec:para-generation}).

    \item For each of these paraphrases, we sample $[4,8]$ texts for a target author, extract their \textsc{Style} embeddings and mean pool the results.

    \item Using these $5$ pairs of (paraphrase, \textsc{Style} embedding), we sample $5$ corresponding outputs from \textsc{TinyStyler}$_\textsc{recon}$ with $\texttt{top-p}=0.80$ and $\tau=1.0$.

\end{enumerate}

\subsubsection{Filtering}

After generating $5$ candidate outputs for each of our $160$k author pairs, we filter low quality outputs.

\begin{enumerate}
    \item First, we filter all candidates that are identical to their input, and those with hallucinated links, which we filter with regex rules.
    \item Next, we filter all candidates that have low meaning preservation scores. We use MIS \cite{babakov-etal-2022-large} and SimCSE \cite{gao2022simcse}, and normalize SimCSE scores between $[0,1]$. After reviewing example outputs for errors and hallucinations, we selected $0.7$ as the threshold for both models.
    \item After filtering candidates with low automatic meaning preservation scores, we filter the remaining outputs with low transfer accuracy scores. We compute \textit{Away} and \textit{Towards} metrics using \textsc{Style} Embeddings. We filter outputs that have an $\textit{Away}<0.9$ and $\textit{Towards}<0.30$.
    \item Finally, when multiple candidates remain for a given source text, we select the output that maximizes $G(G(Away,Towards), Sim)$. Here,  unlike \citet{patel2022lowresource}, we skip normalizing by $MIS(source, target)$ when computing \textit{Sim}.
\end{enumerate}

The resulting data comprises approximately $40$k high quality examples.

\subsubsection{Fine-tuning for \textsc{TinyStyler}}
\label{sec:self-distill-ft}

We resume training our model from Appendix \ref{sec:style-recon-training}, on the resulting High Quality Dataset. Unlike the previous train, we skip paraphrasing and condition directly on the source text, along with all texts from a target author. We otherwise use the same hyperparameters in Table \ref{table:hparam}. We select the checkpoint with the lowest validation loss, which occurred after $20000$ steps.

\section{Evaluation Details}

\subsection{Authorship Transfer}
\label{sec:authorship-eval}

We evaluate authorship transfer on the dataset introduced by \citet{patel2022lowresource}. We evaluate on their three dataset splits:

\begin{itemize}
    \item \textit{Random}: Random source and target authors.
     \item \textit{Single}: All posts belong to a popular college football subreddit.
      \item \textit{Diverse}: Source and target authors with posts on diverse topics across $13$ or more different subreddits. 
    
\end{itemize}

Each split contains $15$ Reddit source authors and $15$ Reddit target authors. We ensure that all authors in these evaluation sets are excluded from the training data.

When performing authorship style transfer with \textsc{TinyStyler}$_{\textsc{recon,rerank(5)}}$, we sample $5$ paraphrases. Then, for each paraphrase, we sample $8$ texts by the target author, and extract \textsc{Style} embeddings from these. We then generate an output for each (paraphrase, \textsc{Style} embedding) pair, and select the output with the highest value of $G(G(Away,Towards), Sim)$. We compute these metrics using \textsc{Style} embeddings \cite{wegmann-etal-2022-author}, rather than \textsc{UAR} \cite{rivera-soto-etal-2021-learning}. For \textsc{TinyStyler}$_{\textsc{recon}}$, we select the first result.

For \textsc{TinyStyler}, we condition on the source text and all $16$ examples for the target author. When additionally performing ranking for \textsc{TinyStyler}$_{\textsc{rerank(5)}}$, we sample $5$ outputs and again select the result with the highest aggregate score.

\subsection{Formality Transfer}

To evaluate formality transfer, we consider the Entertainment and Music subset of the GYAFC dataset \cite{rao-tetreault-2018-dear}. For all approaches, we consider the $1082$ original formal examples and the $1416$ original informal examples.

To measure \textit{Accuracy}, we use a holdout off-the-shelf model,\footnote{\url{https://huggingface.co/s-nlp/xlmr_formality_classifier}} trained on the XFormal Corpus \cite{dementieva-etal-2023-detecting, briakou2021xformal}. For \textit{Fluency}, we use a model\footnote{\url{https://huggingface.co/textattack/roberta-base-CoLA}} trained on the CoLA dataset \cite{morris2020textattack, warstadt-etal-2019-neural}.

For our internal Formality classifier, we fine-tune a \texttt{roberta-base} \cite{liu2019roberta} model with the hyperparameters in Table \ref{table:hparam3} on $85\%$ of the GYAFC Entertainment Music training set, and validate on the remaining training samples. 

\begin{table}[h!]
\centering
\begin{tabular}{@{}lc@{}}
\toprule
\textbf{Hyperparameter} & \textbf{Value} \\ \midrule
Pretrained Ckpt & \texttt{\small roberta-base} \\
Learning Rate & $5 \times 10^{-5}$ \\
Batch Size & 128 \\
Optimizer & Adam \\
Weight Decay & 0.01 \\
Schedule & Constant \\
Total Steps & 2700 \\ \bottomrule
\end{tabular}
\caption{Fine-tuning hyperparameters for the Internal formality classifier used for \textsc{Mix \& Match} and \textsc{ParaGuide} baselines.}
\label{table:hparam3}
\end{table}

We use our internal classifier for our \textsc{ParaGuide} and \textsc{M\&M} approaches (See Appendix \ref{sec:baselines}). We also use our internal classifier to determine high probability examples of each class (with probability $>0.95$) to serve as exemplars for the LLM methods and \textsc{TinyStyler} models. We randomly select $128$ of these formal and informal exemplars from the Tune set of GYAFC. For \textsc{TinyStyler}, along with \textsc{Gpt-4}, and \textsc{Gpt-3.5}, we sample $16$ of these to use as examples for each inference. We sample $64$ per inference for \textsc{TinyStyler}$_{\textsc{ex}=64}$.

\subsection{Additional Baseline Details}
\label{sec:baselines}

For our authorship evaluations, we use the baselines from \citet{patel2022lowresource}. Additionally, we reproduce \textsc{ParaGuide} by fine-tuning the publicly available SSD-LM \cite{li2022diffusionlm} checkpoint\footnote{\url{https://huggingface.co/xhan77/ssdlm}} on the training dataset described in Appendix \ref{sec:dataset-generation}. We extend the input and output token lengths to $80$, but otherwise use the original paper's hyperparameters.

\begin{table}[h!]
\centering
\begin{tabular}{@{}lc@{}}
\toprule
\textbf{Hyperparameter} & \textbf{Value} \\ \midrule
Pretrained Ckpt & \texttt{\small xhan77/ssdlm} \\
Learning Rate & $5 \times 10^{-6}$ \\
Batch Size & 64 \\
Grad Accum. & 2 \\
Optimizer & Adam \\
Weight Decay & 0.01 \\
Schedule & Constant \\
Diffusion Steps & 200 \\
Context Size & 80 \\
Output Size & 80 \\
Warm-up Steps & 2000 \\
Total Steps & 2000000 \\ \bottomrule
\end{tabular}
\caption{Fine-tuning hyperparameters for the \textsc{ParaGuide} baseline}.
\label{table:hparam2}
\end{table}

For \textsc{ParaGuide}, we perform authorship transfer using style guidance from \textsc{Style} embeddings \cite{wegmann-etal-2022-author, horvitz2024paraguide}. For \textsc{Mix and Match} \cite{mireshghallah2022mix} we use the hyperparameters for the \textit{Hamming} and \textit{Disc} configurations from the original paper for the formality transfer task. Additionally, we use \texttt{RoBERTA-Large} \cite{liu2019roberta} as the base language model.  For both \textsc{ParaGuide} and \textsc{Mix \& Match} we use our internal classifier trained on the training subset of GYAFC \cite{rao-tetreault-2018-dear}.  We estimate a total compute budget of $600$ GPU hours for all \textsc{TinyStyler} and baseline experiments.

Finally, we also prompt \textsc{Gpt-3.5} (\texttt{gpt-3.5-turbo-0125}) and \textsc{Gpt-4} (\texttt{gpt-4-turbo}) for both authorship and formality transfer, using the prompts included in Appendix \ref{sec:prompts}.

\section{Human Evaluation}
\label{sec:human_eval}
For our human evaluation, we generate outputs for  $150$ examples ($75 \rightarrow \textit{Informal}$, $75 \rightarrow \textit{Formal}$) for each approach. We recruited $11$ English speakers, all of whom were graduate student volunteers. We divided the annotations among these speakers, assigning three annotators per example. We assign labels based on majority vote. Our instructions to human annotators are included in Appendix \ref{sec:instructions}.

We measure inter-annotator agreement with Krippendorff's $\alpha$:

\begin{table}[h]
\centering
\begin{tabular}{lcc}
\toprule
\textbf{Label} & \textbf{Krippendorff's $\alpha$} \\
\midrule
Meaning Preservation & 0.55 \\
Fluency & 0.58 \\
Formality & 0.61 \\

\bottomrule
\end{tabular}
\caption{Inter-annotator agreement, computed with Krippendorff's $\alpha$.}

\label{krip}
\end{table}

\section{Timing}
\label{sec:timing}

In Table \ref{tab:timing}, we report timing results on $200$ examples from the Formal $\rightarrow$ Informal transfer task. To estimate timing information, we generate outputs for $200$ samples. For each local approach, we perform inference on an NVIDIA-A100 GPU.

\begin{table}[h]
    \centering
   \begin{tabular}{lc}
\toprule
Method & Seconds/Iter \\
\midrule
\multicolumn{2}{l}{\textit{Large Language Models}} \\
\midrule
\textsc{Gpt-3.5} & 0.70  \\
\textsc{Gpt-4} & 1.56 \\
\midrule
\multicolumn{2}{l}{\textit{Controllable Text Generation}} \\
\cmidrule{1-2}
$\textsc{M\&M}_{\textsc{Disc}}$ & 69.3 \\
$\textsc{M\&M}_{\textsc{Ham}}$ & 68.2  \\
$\textsc{PGuide}_{\lambda=200}$ & 17.72 \\
$\textsc{TStyler}$ & 0.47 \\
$\textsc{TStyler}_{\textsc{ex}=64}$ & 0.43  \\
\bottomrule
\end{tabular}
    \caption{Timing information on a sample of the  Formal $\rightarrow$ Informal task ($n=200$).}
    \label{tab:timing}
\end{table}

\section{Prompts}
\label{sec:prompts}

We prompt \textsc{Gpt-3.5} and \textsc{Gpt-4} for all tasks using OpenAI's chat completions API.\footnote{\url{https://platform.openai.com/docs/guides/text-generation/chat-completions-api}}

\subsection{Authorship Transfer}
    
   \begin{tcolorbox}[breakable, enhanced]

\begin{lstlisting}

message='The following comments are written by a single author: \n'
for i, text in enumerate(examples):
    message += json.dumps({'text':text})+'\n'
message += "\n\nCan you rewrite the following comment to make it look like the above author's style:\n"
message += json.dumps({'text':original_text})+'\n' 

client.chat.completions.create(
    model=model_name,
    response_format={ "type": "json_object" },
    messages=[
    {"role": "system", "content": "You are a helpful assistant designed to output JSON."},
    {"role": "user", "content": message}
    ]

\end{lstlisting}

\end{tcolorbox}

\subsection{Formality Transfer}
\begin{tcolorbox}[breakable, enhanced]
\begin{lstlisting}
message = f'The following texts are written in {target_style} style: \n'
for i, text in enumerate(examples):
    message += json.dumps({'text':text})+'\n'
message += f"\n\nCan you rewrite the following text to make it look like the above {target_style} style:\n"
message += json.dumps({'text':original_text})+'\n'

client.chat.completions.create(
    model=model_name,
    response_format={ "type": "json_object" },
    messages=[
        {"role": "system", "content": "You are a helpful assistant designed to output JSON."},
        {"role": "user", "content": message}
    ]
    
\end{lstlisting}
\end{tcolorbox}

\section{Human Evaluation Instructions}
\label{sec:instructions}

\textbf{Note:} These examples are selected directly from GYAFC \cite{rao-tetreault-2018-dear}, which contains offensive content.

\lstset{frame=tb,
  language=python,
  aboveskip=3mm,
  belowskip=3mm,
  showstringspaces=false,
  columns=flexible,
  basicstyle={\small\ttfamily},
  numbers=none,
  numberstyle=\color{black},
  keywordstyle=\color{black},
  commentstyle=\color{black},
  stringstyle=\color{black},
  breaklines=true,
  breakatwhitespace=true,
  tabsize=3
}

\begin{tcolorbox}[breakable, enhanced]
\textbf{Instructions:}
\begin{lstlisting}
Each annotator has been assigned a series of very short texts to review. Each example consists of a reference and output text.
We would like you to evaluate the output text across three criteria:
1) Similarity to the reference. Is the meaning of the reference preserved by the output? (0=No, 1=Yes)
2) Well-formedness/Fluency. Does the output look like a text that could reasonably appear on an internet forum? Is it a coherent? (0=Badly-Formed, 1=Well-formed)
3) Formality. Is the output text informal or formal? (0=informal, 1=formal)

Empty outputs can be marked with all 0s.
Please avoid consulting other annotators/annotations.
\end{lstlisting}
\textbf{Examples of formal text:}
\begin{lstlisting}
I like Rhythm and Blue music.
There's nothing he needs to change.
It does not exist.
Mine is book by Steve Martin called 'The Pleasure of my Company'.
What differentiates a mosquitoo from a blonde?
They're pretty good. Also, that's a good song.
I do not think Beyonce can sing, dance, or act. You mentioned Rihanna, who is that?
I was unaware that you were in law enforcement, as well.
I called to say 'I Love You
I would most likely not vote for him, although I believe Melania would be the most attractive First Lady in our country's history.
\end{lstlisting}
\textbf{Examples of informal text:}
\begin{lstlisting}
Is Any Baby Really A Freak.
aspen colorado has he best music festivals, you sit all over the moutians its  on and just hang out
You can get almost anything on ebay!
everybody is Dying to get in
not idiots like 50 cent and his whole Gay unit.those kinds of ppl give hip-hop a bad name.
different from what I've seen though
I want to be on TV!
dont let anyone decide the fate but you.
50 is just riding coattails with that movie.
The blind klan guy is hilarious!
\end{lstlisting}
\end{tcolorbox}

\end{document}